\documentclass[letterpaper,10pt,conference]{ieeeconf}
\IEEEoverridecommandlockouts
\usepackage{graphicx}
\usepackage{amsmath}
\usepackage{amssymb}
\usepackage{booktabs}
\usepackage{hyperref}
\usepackage{cite}
\usepackage{xcolor}
\usepackage{array}

\title{\LARGE \bf HorizonNav: Geometry-Guided Video Diffusion for Anticipatory 3D Robot Navigation}

\author{Hajira Amjad, Valerii Serpiva, Viacheslav Liubimov, Didar Seyidov and Dzmitry Tsetserukou
\thanks{The authors are with the Intelligent Space Robotics Laboratory, Skolkovo Institute of Science and Technology Moscow, Bolshoy Boulevard 30, bld. 1, 121205, Moscow, Russia.
\tt \{Hajira.Amjad, Valerii.Serpiva, V.Lyubimov, Didar.Seyidov, D.Tsetserukou\}@skoltech.ru}}

\begin{document}
\maketitle


\begin{abstract}
We present HorizonNav, a video diffusion-based framework for anticipatory robot navigation that transforms a single RGB image and a text prompt into a geometrically aligned representation of plausible surrounding 3D space. Given the initial observation and a textual description of the reconstruction objective, a vision-language model, guided by designed instructions, interprets the scene and constructs structured prompts for a video diffusion model. The resulting video sequence provides plausible future observations of the environment, which are subsequently processed using monocular ORB-SLAM3 for camera pose estimation and Video Depth Anything for dense depth prediction. Our central contribution is a geometry-guided depth alignment stage that combines the estimated camera poses with triangulated visual landmarks to constrain dense depth predictions across multiple views. This alignment improves the geometric consistency of the reconstructed 3D environment, addressing inconsistencies that temporal coherence of the generated video alone cannot resolve. The experimental evaluation of navigation in imagined 3D environment shows that across a 19-scene horizon study, planning-feasible scenes increase from 13/19 at P20 to 15/19 at P81, while successful RRT trials increase from 62/90 to 75/90. In a separate corridor study, 6 of 25 evaluable generated-camera trajectories violate the reconstructed collision constraints; RRT planner recovers feasible alternatives in 5 of those 6 cases. A larger archive contains 58 usable reconstructions from 67 attempted cases, of which 36 support a downstream plan and 22 do not, exposing reconstruction and free-space failure modes rather than hiding them without lidar
Across 15 corridor reconstructions, the mean reconstruction error is approximately 0.30 m at matched structural landmarks.
\end{abstract}


\section{Introduction}

Autonomous navigation in unfamiliar environments requires robots to act before their surroundings are fully explored. Visual simultaneous localization and mapping (SLAM) reconstructs observed scene structure, while regions outside the current field of view remain unknown. A robot that can form plausible spatial hypotheses before motion may be able to evaluate candidate routes earlier, but such hypotheses are useful only if their geometry is treated cautiously rather than as ground truth.

Recent generative world models use predicted observations for navigation and action reasoning. Navigation World Models~\cite{navwm2026} predict action-conditioned observations, while ImagiNav~\cite{imagenav2603} combines language-guided video generation with trajectory extraction. WonderWorld~\cite{yu2025wonderworld}, WorldExplorer~\cite{schneider2025worldexplorer}, and GEN3C~\cite{gen3c2025} further demonstrate that generative models can expose spatial structure beyond a single image. However, a generated trajectory need not be safe to execute, and generated views may contain deformations or inconsistent occlusions that alter free-space connectivity.

Dense depth estimation provides per-pixel depth information for generated observations, but it does not by itself enforce a geometrically aligned 3D reconstruction across views. Video Depth Anything [6] produces temporally coherent depth maps and explicitly achieves this without introducing geometric priors or video generation priors. However, temporal consistency alone does not necessarily guarantee accurate multi-view geometry. In contrast, methods such as VGGT [7] and Depth Anything 3 [8] explicitly target multi-view geometric estimation, while World Reconstruction From Inconsistent Views [9] addresses contradictions between generated observations through non-rigid alignment. These developments motivate combining exploratory video generation with explicit geometric constraints to transform temporally consistent depth predictions into spatial representations that are more geometrically coherent and useful for robot planning.

We present HorizonNav a framework that transforms a single RGB image and a language prompt into a geometrically aligned representation of plausible surrounding space. Starting from one RGB observation and a language task, a vision-language model produces structured exploration instructions and Wan~2.2 generates complementary virtual camera motion. ORB-SLAM3~\cite{campos2021orbslam3} estimates the corresponding virtual camera poses and sparse 3D landmarks, whereas Video Depth Anything predicts dense relative depth maps for the generated frames. We align the dense depth estimates with the sparse geometric structure recovered by ORB-SLAM3, jointly resolving scale and geometric consistency across views. A measured scene reference resolves monocular scale, after which RRT~\cite{lavalle1998rapidly} searches the resulting obstacle representation independently of the generated camera motion. HorizonNav combines exploration-oriented VLM prompting, sparse-landmark-guided depth alignment, and multivideo map fusion to connect generative visual exploration with explicit spatial reasoning for navigation.

This design deliberately separates three questions that are often conflated: whether generated frames contain enough repeatable visual structure for multi-view reconstruction, whether the resulting spatial hypothesis preserves free-space connectivity well enough for planning, and whether a map-feasible plan transfers to the physical robot.





\begin{figure*}[t]
  \centering
  \includegraphics[width=0.95\textwidth]{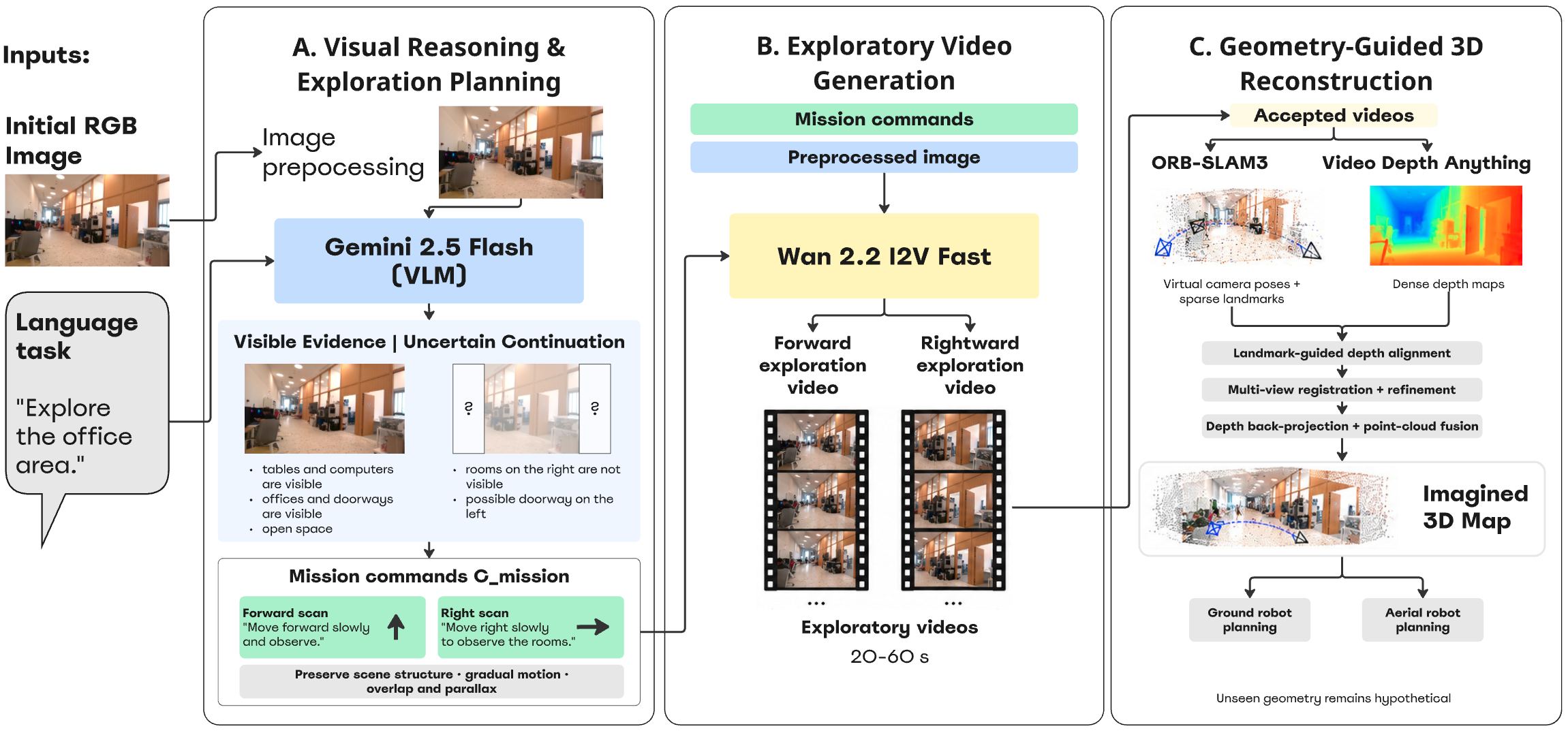}
  \caption{Overview of the HorizonNav framework. Visual reasoning and exploration planning: a VLM analyzes the initial RGB image and language task to generate structured exploration commands. Exploratory video generation: Wan 2.2 synthesizes plausible future observations along multiple virtual camera trajectories. Geometry-guided 3D reconstruction: ORB-SLAM3 estimates camera poses and sparse landmarks, while Video Depth Anything predicts dense depth maps. Landmark-guided depth alignment, multi-view registration, and point-cloud fusion produce a geometrically consistent imagined 3D map for subsequent robot navigation and path planning.}
  \label{fig:system_architecture}
\end{figure*}

\section{Related Work}

\subsection{Generated Futures for Robot Planning}

Generated video is increasingly used to reason about robot motion before execution. Navigation World Models~\cite{navwm2026}, ImagiNav~\cite{imagenav2603}, and NavDreamer~\cite{huang2026navdreamer} use predicted observations to evaluate candidate behavior or derive actions and waypoints. Schr\"{o}dinger's Navigator~\cite{he2025schrodinger}, Aerial Navigation World Model~\cite{zhang2026anwm}, Uncertainty-Aware Navigation World Model~\cite{zhu2026uanwm}, and GVP-WM~\cite{ziakas2026gvpwm} extend this idea by reasoning over multiple futures, uncertainty, or learned feasibility constraints.

These methods show that generated futures can be useful for navigation; in many cases, the final decision remains tied to the generated sequence or to a learned world representation. HorizonNav studies a different use of generated video. The generated video serves as a form of visual imagination, enabling the reconstruction of a plausible 3D representation of the surrounding environment, which is subsequently used for robot navigation and path planning.

\subsection{Geometric Reconstruction from Generated Observations}

Several recent systems connect generative models with explicit 3D structure. Dream-SLAM~\cite{dreamslam2026} introduces dreamed observations into an active SLAM loop, while WorldMAP~\cite{chen2026worldmap} converts generated futures into persistent spatial representations. WonderWorld~\cite{yu2025wonderworld} and WorldExplorer~\cite{schneider2025worldexplorer} build explorable 3D environments from generated content. Voyager~\cite{huang2025voyager}, Sora3R~\cite{mai2025sora3r}, Geometry Forcing~\cite{wu2025geometryforcing}, and GEN3C~\cite{gen3c2025} improve geometric or camera consistency by changing the generation process itself.


Direct multi-view geometry methods provide another relevant comparison. DUSt3R~\cite{wang2024dust3r}, MASt3R~\cite{leroy2024mast3r}, VGGT~\cite{wang2025vggt}, Depth Anything 3~\cite{lin2026depthanything3}, and MASt3R-Nav~\cite{masternav2605} estimate geometry directly from visual observations. World Reconstruction From Inconsistent Views~\cite{hoellein2026world} specifically addresses contradictions in generated sequences through non-rigid alignment. HorizonNav instead asks whether generated views can provide enough trackable multi-view evidence for a planning-oriented spatial hypothesis without modifying the generator.

\subsection{3D Reconstruction and Navigation Planning}

Single-image depth methods such as MiDaS~\cite{ranftl2020midas}, Depth Anything V2~\cite{yang2024depthanything}, Depth Pro~\cite{bochkovskii2024depthpro}, and MoGe~\cite{moge2024} provide dense geometric cues from individual images. Video Depth Anything~\cite{chen2025videodepthanything} improves temporal consistency across video frames. In HorizonNav, dense depth is not assumed to be geometrically correct merely because it is temporally smooth; we evaluate VDA separately as a densification diagnostic.

ORB-SLAM3~\cite{campos2021orbslam3} provides the primary sparse multi-view geometry used in the reported planning experiments. RRT~\cite{lavalle1998rapidly} is the downstream planner. Neither ORB-SLAM3 nor RRT is presented as an algorithmic contribution. The research question is whether generated virtual observations can supply enough geometric evidence to construct a useful pre-motion spatial hypothesis, while keeping the executed plan independent of the generator's camera trajectory.

\section{System Architecture} 
\label{sec:method}

As presented in Fig.~\ref{fig:system_architecture} consists of three
modules: visual reasoning and virtual exploration planning,
video generation, and geometry-guided 3D reconstruction.

Given an initial image $I_{\mathrm{input}}$ and a user task
$C_{\mathrm{task}}$, a vision-language model (VLM), Gemini 2.5 Flash \cite{deepmind2024gemini}, analyses visible scene elements, including
landmarks, obstacles, traversable regions, and possible occlusions. It produces a grounded
exploration command
\begin{equation}
C_{\mathrm{explore}} =
f_{\mathrm{VLM}}(I_{\mathrm{input}}, C_{\mathrm{task}}),
\end{equation}
which specifies the target direction, feasible motion primitives, visual
anchors that should remain consistent with the input image, and regions
that require further observation. Prompt constraints instruct the model
to distinguish visible evidence from uncertain areas and avoid introducing
unjustified objects or scene changes. Thus, the VLM converts a high-level
mission into an exploration-oriented description that guides the
generation process while reducing semantic hallucinations.

The standardized initial frame $I_{\mathrm{std}}$ and the exploration
command $C_{\mathrm{explore}}$ condition the Wan 2.2 I2V Fast diffusion
model \cite{wan2025wan}, which generates candidate FPV videos:
\begin{equation}
V_{\mathrm{gen}} =
f_{\mathrm{I2V}}(I_{\mathrm{std}}, C_{\mathrm{explore}}).
\end{equation}
Each generated sequence represents a continuation of the
current observation along the planned exploratory motion, for example,
moving forward through a visible opening or observing an occluded region
from a complementary viewpoint. The initial image and VLM-derived visual
anchors constrain the generation, whereas multiple candidate sequences
can be sampled when the unobserved space is ambiguous.

The generated video is converted into a spatial representation by a
geometry reconstruction module. ORB-SLAM3 \cite{campos2021orbslam3} estimates
the camera trajectory $\{\mathbf{T}_{cw}^{t}\}_{t=1}^{T}$ and a sparse
set of triangulated 3D landmarks $\mathcal{M}$. In parallel, Video Depth
Anything predicts a dense relative depth map $\hat{D}_t$ for each frame.
The sparse SLAM landmarks are used to align the relative depth maps with
the trajectory coordinate frame, producing corrected dense maps
$D_t^{*}$. The corrected maps are back-projected and transformed using
the estimated camera poses to form a unified point cloud
$\mathcal{P}_{\mathrm{3D}}$.

\subsection{Trajectory-Guided Alignment of Relative Depth and Sparse SLAM Landmarks}
\label{sec:depth_slam_alignment}

Given a difution generated video sequence $\{I_t\}_{t=1}^{T}$, ORB-SLAM3
provides the camera poses by monocular visual odometry $\mathbf{T}_{cw}^{t} \in SE(3)$ and a sparse set of triangulated map points $\mathcal{M}=\{\mathbf{P}_j^w\}$.
Here, $\mathbf{T}_{cw}^{t}$ transforms a point from the world frame to the camera frame. For each frame, Video Depth Anything predicts a dense relative depth map $\hat{D}_t:\Omega\rightarrow \mathbb{R}^{+}$. Although the predicted maps are temporally consistent, their scale and offset are not necessarily aligned with the geometric coordinate system estimated by SLAM. Therefore, we use the sparse ORB-SLAM3 landmarks as geometric anchors for depth calibration.

For each map point $\mathbf{P}_j^w$ observed in frame $t$, its
coordinates in the camera frame are computed as
\begin{equation}
\begin{bmatrix}
\mathbf{p}_{j,t}^{c} \\
1
\end{bmatrix}
=
\mathbf{T}_{cw}^{t}
\begin{bmatrix}
\mathbf{P}_j^w \\
1
\end{bmatrix},
\qquad
z_{j,t}
=
\mathbf{e}_3^{\top}\mathbf{p}_{j,t}^{c},
\label{eq:slam_depth}
\end{equation}
where $\mathbf{e}_3=[0,0,1]^\top$ and $z_{j,t}$ is the metric depth of
the landmark along the optical axis in the SLAM coordinate system.
Using the camera intrinsic matrix $\mathbf{K}$, the corresponding image
location is
\begin{equation}
\mathbf{u}_{j,t}
=
\pi\left(\mathbf{K}\mathbf{p}_{j,t}^{c}\right),
\label{eq:projection}
\end{equation}
where $\pi(\cdot)$ denotes perspective division.

Only landmarks with positive depth, valid image projection, low
reprojection error, and a sufficient number of observations are retained.
The relative depth value at each landmark position is obtained by
bilinear interpolation:
\begin{equation}
\hat{d}_{j,t}
=
\hat{D}_t(\mathbf{u}_{j,t}).
\label{eq:depth_sampling}
\end{equation}

\subsubsection{Robust scale and shift estimation}

We model the relation between the relative depth prediction and the
SLAM-based depth using an affine transformation:
\begin{equation}
\tilde{D}_t(\mathbf{u})
=
\alpha_t \hat{D}_t(\mathbf{u})+\beta_t,
\qquad
\alpha_t > 0,
\label{eq:affine_depth}
\end{equation}
where $\alpha_t$ and $\beta_t$ are the scale and shift parameters,
respectively. They are estimated from the sparse landmark correspondences
by minimizing a robust reprojection-depth error:
\begin{equation}
\begin{aligned}
\min_{\{\alpha_t,\beta_t\}} \;&
\sum_{t=1}^{T}\sum_{j\in\mathcal{V}_t}
w_{j,t}\rho\left(\alpha_t \hat{d}_{j,t}+\beta_t-z_{j,t}\right) \\
&+ \lambda_{\alpha}\sum_{t=2}^{T}(\alpha_t-\alpha_{t-1})^2
+ \lambda_{\beta}\sum_{t=2}^{T}(\beta_t-\beta_{t-1})^2 .
\end{aligned}
\label{eq:depth_alignment}
\end{equation}

where $\mathcal{V}_t$ is the set of valid ORB-SLAM3 landmarks in frame
$t$, $w_{j,t}$ is an optional landmark confidence weight, and
$\rho(\cdot)$ is the Huber loss. The temporal regularization terms
prevent frame-wise scale fluctuations while allowing gradual changes
caused by video-generation artifacts.

\subsubsection{Local residual correction}

A global scale and shift correct low-frequency depth errors but may not
fully align local structures. Therefore, sparse residuals at landmark
locations are computed as
\begin{equation}
r_{j,t}
=
z_{j,t}-\tilde{D}_t(\mathbf{u}_{j,t}).
\label{eq:sparse_residual}
\end{equation}
These residuals are propagated to nearby pixels using an edge-aware
kernel:
\begin{equation}
\Delta D_t(\mathbf{u})
=
\frac{
\sum_{j\in\mathcal{V}_t}
k(\mathbf{u},\mathbf{u}_{j,t})\,r_{j,t}
}{
\sum_{j\in\mathcal{V}_t}
k(\mathbf{u},\mathbf{u}_{j,t})+\epsilon
},
\label{eq:residual_field}
\end{equation}
with
\begin{equation}
k(\mathbf{u},\mathbf{u}_{j,t})
=
\exp\left(
-\frac{\|\mathbf{u}-\mathbf{u}_{j,t}\|_2^2}{2\sigma_p^2}
-\frac{\|I_t(\mathbf{u})-I_t(\mathbf{u}_{j,t})\|_2^2}{2\sigma_I^2}
\right).
\label{eq:edge_aware_kernel}
\end{equation}
The final corrected depth map is
\begin{equation}
D_t^{*}(\mathbf{u})
=
\max\left(
\epsilon,\,
\tilde{D}_t(\mathbf{u})+\Delta D_t(\mathbf{u})
\right).
\label{eq:corrected_depth}
\end{equation}
The image-dependent term in Eq.~\eqref{eq:edge_aware_kernel} limits
residual propagation across object boundaries and preserves depth
discontinuities.

\subsubsection{Dense point cloud reconstruction and fusion}

For each valid pixel $\mathbf{u}=[u,v,1]^\top$, the corrected depth is
back-projected into the camera frame:
\begin{equation}
\mathbf{X}_{t}^{c}(\mathbf{u})
=
D_t^{*}(\mathbf{u})
\mathbf{K}^{-1}
\begin{bmatrix}
u\\v\\1
\end{bmatrix}.
\label{eq:backprojection}
\end{equation}
The point is then transformed into the common world frame using the
inverse ORB-SLAM3 pose:
\begin{equation}
\mathbf{X}_{t}^{w}(\mathbf{u})
=
\mathbf{R}_{wc}^{t}\mathbf{X}_{t}^{c}(\mathbf{u})
+
\mathbf{t}_{wc}^{t},
\qquad
\mathbf{T}_{wc}^{t}
=
\left(\mathbf{T}_{cw}^{t}\right)^{-1}.
\label{eq:world_transform}
\end{equation}

Finally, dense point clouds from all frames are merged and filtered with
voxel-grid downsampling:
\begin{equation}
\mathcal{P}_{\mathrm{dense}}
=
\operatorname{VoxelFuse}
\left(
\bigcup_{t=1}^{T}
\left\{
\mathbf{X}_{t}^{w}(\mathbf{u})
\mid
\mathbf{u}\in\Omega
\right\}
\right).
\label{eq:cloud_fusion}
\end{equation}

\subsection{RRT-Based Navigation in the Reconstructed 3D Space}

We employ the Rapidly-exploring Random Tree (RRT) algorithm for planning
collision-free paths in the reconstructed 3D representation. The point
cloud $\mathcal{P}_{\mathrm{3D}}$ is first converted into an occupancy
map, in which observed obstacles are inflated by the robot radius and a
safety margin. This produces a conservative free configuration space
$\mathcal{X}_{\mathrm{free}}$ for navigation.

RRT is well suited to this task because the reconstructed environment can
contain irregular obstacle boundaries, narrow passages, and incomplete
geometry. In contrast to gradient-based trajectory optimization methods,
RRT does not require a differentiable map or an accurate analytical model
of the environment. It iteratively samples candidate configurations,
connects them to the nearest node of the search tree, and accepts an
extension only when the corresponding motion segment is collision-free.

For a ground robot, the sampled configuration can be represented as
\begin{equation}
\mathbf{q}_{\mathrm{ground}} = [x,\, y,\, \psi]^{\top},
\end{equation}
whereas for a UAV the planner operates directly in 3D:
\begin{equation}
\mathbf{q}_{\mathrm{UAV}} = [x,\, y,\, z,\, \psi]^{\top}.
\end{equation}
Therefore, the same planning framework can be applied to both platforms,
while accounting for the additional altitude dimension of aerial motion.

The primary objective is to quickly obtain a feasible and safe path in a
partially known environment rather than to find the globally shortest
trajectory. This property is particularly useful when the 3D hypothesis
is updated after new observations become available: the planner can be
executed again from the current robot state using the updated occupancy
map. The resulting RRT path can subsequently be shortened and smoothed
before execution. 

Since the reconstructed scene is a geometrically grounded hypothesis
rather than an exact map of unseen space, obstacle inflation and online
perception remain necessary during execution. Thus, RRT provides a
practical planning layer that converts the imagined 3D representation
into conservative, executable navigation trajectories.

\begin{figure*}[t]
    \centering
    \includegraphics[width=0.8\textwidth]{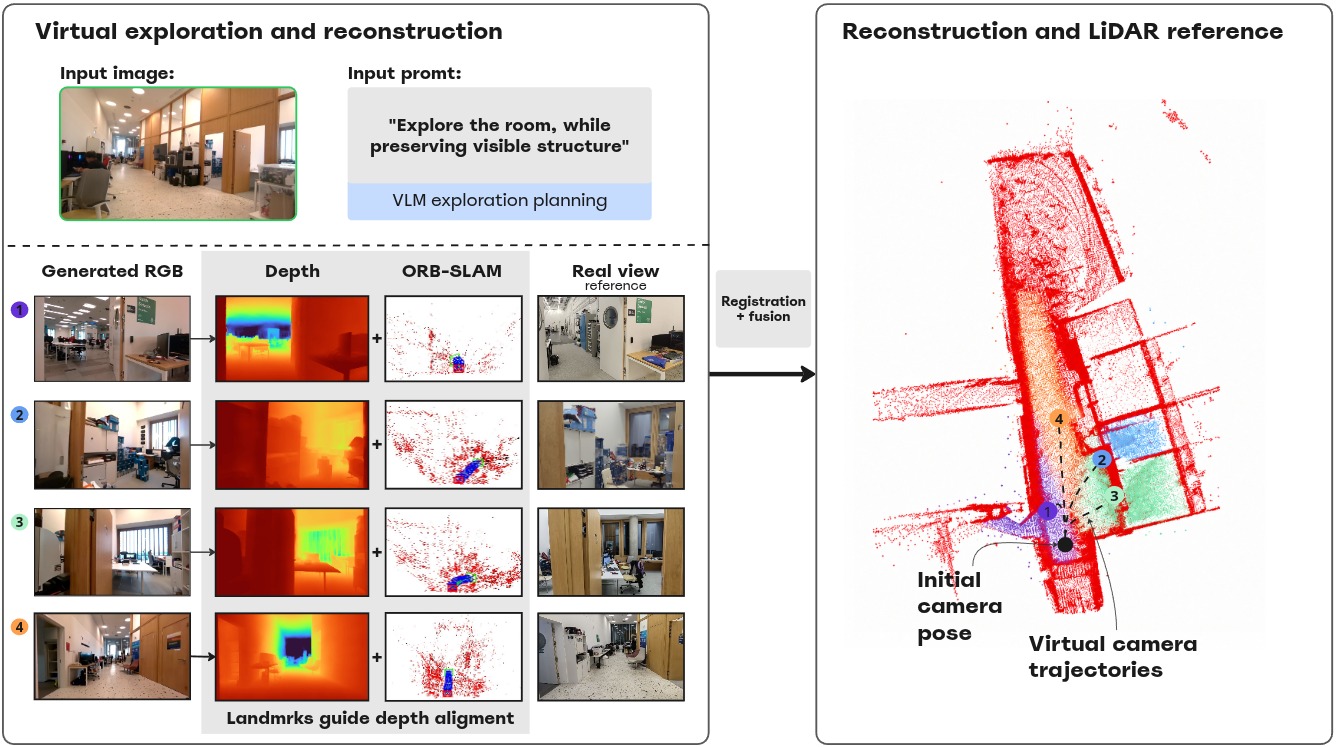}
    \caption{Qualitative generated-view geometry. Left: representative exploration branches showing generated RGB, dense-depth diagnostic, ORB-SLAM3 sparse structure, and corresponding real-view references. Right: reconstructed structure registered to the external LiDAR reference for visualization and evaluation; LiDAR is not available to HorizonNav during reconstruction or planning.}
    \label{fig:reconstruction}
\end{figure*}

\section{Experimental Evaluation}

The evaluation is organized around the main scientific question of HorizonNav: whether generated virtual views can provide geometric evidence that is useful for independent pre-motion planning. We evaluate four aspects separately: planning utility as more generated viewpoints are incorporated, independence between generated camera motion and the executed plan, external geometric diagnostics of the recovered spatial hypothesis, and transfer of a precomputed path to a physical robot. Unless explicitly stated otherwise, feasibility is defined with respect to the reconstructed obstacle representation; it is not treated as proof that unseen physical geometry has been recovered correctly.

\subsection{Planning Utility Across Virtual-View Horizons}

We evaluate generated-view prefixes at P20, P40, P60, and P81. For each horizon, we report (i) the number of scenes for which the reconstructed spatial hypothesis admits at least one RRT path and (ii) seed-level RRT success across the available trials. This experiment measures whether additional generated viewpoints improve downstream planning utility. It does not by itself measure physical reconstruction accuracy.

\subsection{Alternative Virtual Exploration and Independent Planning}

The second experiment tests a central design choice of HorizonNav: generated camera motion supplies virtual observations for reconstruction, but it is not assumed to be an executable robot trajectory.

The corridor study contains two distinct real trigger observations, S5 and S6. For each trigger image, four exploration commands---FORWARD, LEFT, RIGHT, and ENTER RIGHT---are generated under two random seeds. This produces 16 videos. Each video contains 81 frames at 12~fps and is evaluated at P40 and P81, giving 32 branch-prefix cases.

A branch-prefix case is evaluable when the recovered virtual camera trajectory and the independently planned path can both be expressed in the same metric obstacle representation. For each evaluable case, the ORB-SLAM3 camera trajectory is transformed into that representation and checked against the same collision geometry used by RRT. A generated trajectory is considered infeasible when any point along it violates the 0.20~m collision criterion. The corridor planner uses a step size of 0.08~m, a goal bias of 0.45, and at most 15,000 iterations.

Planner robustness is evaluated by rerunning RRT with 20 fixed seeds while holding the generated video, reconstruction, metric scale, and obstacle geometry unchanged. This isolates planner randomness from variation introduced by generation or reconstruction.

\subsection{Geometric Diagnostics}

The third experiment separates two questions that should not be conflated: whether locally reconstructed structure agrees with independently measured geometry at selected features, and whether the recovered global metric scale is consistent.

For a thin-aperture diagnostic, the S06 gate scene is evaluated using the sparse SLAM reconstruction and the corrected dense VDA+SLAM reconstruction. The opening is deliberately sensitive to small depth errors, so this comparison tests whether dense predictions can alter traversability even after timestamp and metric-frame alignment.

For external reference, we use an independently acquired Livox LiDAR map of the corridor. For each available P40 corridor reconstruction, we first register HorizonNav geometry to LiDAR with an $\mathrm{SE}(3)$ transform while keeping recovered scale fixed. We then repeat the registration with a similarity transform that introduces one additional scale parameter. The resulting post-hoc scale indicates how much global rescaling would be required after metric anchoring. Separately, selected structural landmarks---door-frame corners, wall intersections, and corners of fixed objects---are compared after rigid registration. These measurements are diagnostics of matched structure and global scale, not a complete surface-reconstruction benchmark. LiDAR is used only for evaluation and is not available to HorizonNav during reconstruction or planning.

\subsection{Real-World Flight Validation}

The final experiment asks whether a path that is feasible in the reconstructed spatial hypothesis remains executable on the physical robot. This is distinct from map-level planning: physical execution also introduces reconstruction mismatch, vehicle dynamics, controller error, and trajectory tracking error.

We perform six quadrotor flight trials in the gate environment. In every trial, HorizonNav computes the complete waypoint path before takeoff from the generated-view spatial hypothesis. No observation collected during flight is used to modify the reconstruction or replan the path. Motion capture is used only for closed-loop state feedback and for recording the executed trajectory.

A flight is counted as successful when the recorded trajectory crosses the gate plane within the measured aperture. We additionally report lateral gate-crossing error and trajectory tracking error to characterize the difference between the pre-motion reference path and physical execution.


\section{Results}

\subsection{Planning Utility Across Virtual-View Horizons}

Across the 19-scene horizon study, planning feasibility improves overall as more generated observations become available, although the change is not monotonic for every individual reconstruction. Planning-feasible scenes increase from 13/19 at P20 to 15/19 at P81, while successful RRT trials increase from 62/90 to 75/90. Table~\ref{tab:horizon_results} reports the full horizon sweep.

\begin{table}[t]
\centering
\caption{Planning utility across generated-view horizons.}
\label{tab:horizon_results}
\begin{tabular}{ccc}
\toprule
\textbf{Prefix} & \textbf{Feasible scenes} & \textbf{Successful RRT trials} \\
\midrule
P20 & 13/19 & 62/90 \\
P40 & 14/19 & 70/90 \\
P60 & 14/19 & 70/90 \\
P81 & 15/19 & 75/90 \\
\bottomrule
\end{tabular}
\end{table}

These numbers measure utility with respect to the reconstructed obstacle representation; they are not a surface-accuracy metric. The improvement from P20 to P81 indicates that additional virtual viewpoints can expose geometric evidence useful to the downstream planner, while the non-monotonic case-level behavior motivates explicit failure analysis.

\subsection{Reconstruction Yield and Failure Analysis}

A separate archive contains 67 attempted generated-view reconstruction cases. ORB-SLAM3 produces a usable reconstruction in 58 cases, giving an empirical reconstruction yield of 58/67 on generated sequences. This is evidence that feature-based reconstruction is often viable on the generated videos, but it is not a real-versus-generated keypoint-repeatability benchmark.

Of the 58 usable reconstructions, 36 support a feasible downstream plan and 22 do not; thus 37.9\% of usable spatial hypotheses fail at the planning stage. Inspection of the failed cases identifies recurring mechanisms rather than a single planner defect: insufficient or degenerate virtual camera motion limits triangulation; tracking/reconstruction degradation produces sparse or distorted obstacle structure; and erroneous or expanded geometry can close narrow free-space connections. Longer generation can therefore help by adding viewpoints but can also hurt when later imagined geometry expands into the planning corridor. We consequently report reconstruction yield and planning feasibility as separate outcomes.

\subsection{Independent Planning over Generated Futures}

Of the 32 corridor branch-prefix cases, 25 contain the artifacts required to
evaluate both trajectories in the same reconstructed metric map. Among these
25 cases, 6 generated-camera trajectories violate the 0.20~m collision
criterion. Independent RRT recovers a collision-feasible alternative in 5 of
those 6 cases.

This comparison is important because the planner and the generated trajectory
are evaluated against the same obstacle geometry. The result therefore does
not arise from giving RRT a different map. Instead, it shows that the virtual
camera motion can provide useful observations for reconstruction even when
that motion itself should not be executed.

The P40 reruns remain stable across the 20 planner seeds used in the robustness
test. This indicates that the observed feasibility pattern is not explained by
one favorable RRT seed.

\begin{table}[t]
\centering
\caption{Independent-planning evaluation in the corridor study.}
\label{tab:corridor_independent}
\begin{tabular}{lc}
\toprule
\textbf{Measure} & \textbf{Result} \\
\midrule
Generated branch-prefix cases & 32 \\
Cases evaluable in a common map & 25 \\
Generated trajectories violating collision criterion & 6/25 \\
Violating cases recovered by independent RRT & 5/6 \\
\bottomrule
\end{tabular}
\end{table}

\subsection{3D Geometric Diagnostics}

We compare selected reconstructed structure with an independently acquired Livox LiDAR reference map, used exclusively for external evaluation. After metric anchoring, the reconstruction is registered to the reference using a rigid transformation while keeping scale fixed. Structural agreement is evaluated at corresponding door-frame corners, wall intersections, and corners of fixed objects.

For reconstructed landmarks $\mathbf{x}_j$ and reference landmarks $\mathbf{y}_j$, the mean positional discrepancy is
\begin{equation}
E_{\mathrm{mean}} =
\frac{1}{N}\sum_{j=1}^{N}
\left\|R_L\mathbf{x}_j+\mathbf{t}_L-\mathbf{y}_j\right\|_2
\approx 0.30~\mathrm{m},
\end{equation}
where $(R_L,\mathbf{t}_L)$ aligns the reconstruction with the LiDAR coordinate frame. This value characterizes agreement at the selected matched structural features only; it should not be interpreted as complete surface accuracy, reconstruction completeness, or validation of generated geometry in unseen regions.

Separately, we assess recovered global metric scale across 15 available P40 corridor reconstructions. Fixed-scale $\mathrm{SE}(3)$ registration is compared with similarity registration, which introduces an additional scale parameter $s^\star$. Table~\ref{tab:lidar_scale} summarizes the results.

\begin{table}[t]
\centering
\caption{External LiDAR diagnostic of recovered global metric scale.}
\label{tab:lidar_scale}
\begin{tabular}{lc}
\toprule
\textbf{Measure} & \textbf{Value} \\
\midrule
P40 reconstructions & 15 \\
Optimal scale, median & $0.995\times$ \\
Optimal scale, mean & $0.994\times$ \\
Optimal scale, IQR & $0.994$--$0.997\times$ \\
Median $|s^\star-1|$ & $0.5\%$ \\
$\mathrm{SE}(3)$ correspondence RMSE & $0.325$~m \\
Similarity correspondence RMSE & $0.324$~m \\
Median RMSE reduction & $0.3\%$ \\
\bottomrule
\end{tabular}
\end{table}

The optimal scale remains close to unity, and allowing one additional global scale degree of freedom changes correspondence RMSE only slightly, from 0.325~m to 0.324~m. This supports consistency of the global metric scale after anchoring, but it does not establish local surface accuracy or correctness of unseen geometry.

\begin{table}[t]
\centering
\caption{S06 thin-aperture diagnostic comparing sparse SLAM with corrected dense VDA+SLAM geometry.}
\label{tab:s06_depth}
\begin{tabular}{lc}
\toprule
\textbf{Map used for planning} & \textbf{RRT success} \\
\midrule
Sparse SLAM reconstruction & 5/5 \\
Aligned dense reconstruction & 0/5 \\
\bottomrule
\end{tabular}
\end{table}

Table~\ref{tab:s06_depth} exposes a failure mode rather than an accuracy gain: the corrected dense geometry closes an opening that remains traversable in the sparse reconstruction. This shows why dense predictions cannot be accepted as obstacle evidence solely because they are spatially dense or temporally coherent. The experiment does not support a claim that VDA fusion is more accurate than sparse SLAM.

\section{Discussion}

The experiments support a narrower claim than full 3D scene reconstruction: generated virtual observations can provide enough trackable structure for useful pre-motion planning in some scenes. The 58/67 reconstruction yield in the larger archive directly tests the premise that ORB-SLAM3 can operate on generated frames, while the 36/58 downstream planning rate shows that successful reconstruction is not sufficient. Geometry must also preserve the obstacle boundaries and free-space connectivity that matter to the planner.

The independent-planning experiment strengthens this distinction. Among 25 corridor cases evaluable in a common reconstructed metric map, six generated-camera trajectories violate the reconstructed collision constraints, and independent RRT finds alternatives in five of those six cases. The benefit therefore does not come from executing the motion imagined by the video model; generated motion can be geometrically useful as a source of observations even when it is unsuitable as the robot trajectory.

The external diagnostics also limit what can be claimed about reconstruction quality. The LiDAR scale experiment indicates that little additional global rescaling is required after metric anchoring, yet selected matched structural landmarks still show approximately 0.30~m mean discrepancy. In S06, corrected dense VDA+SLAM geometry closes an aperture for which the sparse SLAM representation supports 5/5 RRT trials. Global scale consistency, local surface accuracy, and planning utility are therefore distinct properties.

Five of six physical flights pass through the gate using paths computed before takeoff. This provides limited end-to-end evidence in the tested gate configuration, not statistical validation of general reconstruction accuracy. The present study is also limited to indoor scenes, one evaluated video generator, fixed structured prompting without a prompt ablation, and no protocol-matched single-image-depth baseline. Historical single-image depth diagnostics used different camera/planner settings and are intentionally excluded from quantitative comparison. A controlled single-image baseline, broader surface-level 3D evaluation, additional environments and robot platforms, and online verification against new physical observations are important next steps.

\section{Conclusion}

We presented HorizonNav, a framework that treats generated video as virtual sensing for pre-motion geometric planning from a single RGB observation. Wan~2.2 proposes hypothetical camera views, ORB-SLAM3 reconstructs sparse geometry, a measured reference resolves metric scale, and goal-biased RRT searches the resulting obstacle representation independently of the generated camera trajectory.

Across the 19-scene horizon study, planning-feasible scenes increase from 13/19 to 15/19 and successful RRT trials from 62/90 to 75/90. In the separate corridor study, independent RRT recovers feasible alternatives in five of six cases where generated-camera motion violates the reconstructed collision constraints. A larger 67-case archive yields 58 usable reconstructions, of which 36 support planning and 22 fail, making the principal failure modes explicit. External LiDAR diagnostics show approximately 0.30~m mean discrepancy at selected matched structural landmarks and a median additional global scale factor of $0.995\times$ across 15 P40 reconstructions. Five of six physical quadrotor trials traverse the gate using paths computed before takeoff.

These results support generated virtual views as a source of planning-oriented geometric evidence, not as proof that the true unseen 3D scene has been recovered. The dense-depth diagnostic further shows that adding geometry can reduce, rather than improve, planning utility when local depth errors create false obstacles. Future work should add a protocol-matched single-image baseline, surface-level reconstruction metrics against independent 3D ground truth, generator and prompt ablations, broader robot/environment testing, and online verification that can accept, revise, or reject imagined geometry as real observations become available.


\bibliographystyle{IEEEtran}
\bibliography{references}

\end{document}